\pdfoutput=1

\documentclass[11pt]{article}

\usepackage{naacl2021}

\usepackage{times}
\usepackage{latexsym}

\usepackage[T1]{fontenc}

\usepackage[utf8]{inputenc}

\usepackage{microtype}

\usepackage{booktabs}
\usepackage{url}
\usepackage{times}
\usepackage{latexsym}
\usepackage{hyperref}

\usepackage{amsmath,amssymb}
\usepackage{graphicx}
\usepackage{tabularx}
\usepackage{multirow}
\usepackage{pifont}

\usepackage[toc,page]{appendix}
\usepackage{xcolor}
\usepackage{floatrow}
\usepackage{makecell}

\newfloatcommand{capbtabbox}{table}[][\FBwidth]

\usepackage{arydshln}

\usepackage{minibox}
\usepackage{tabularx,ragged2e}
\usepackage{tabularx} 
\usepackage{algorithm}
\usepackage[noend]{algpseudocode}
\newcommand\numberthis{\addtocounter{equation}{1}\tag{\theequation}}

\usepackage{bbm}

\newcommand{\ours}{\text{MetricWSD}}
\newcommand{\bert}[1]{{\tt\MakeUppercase{#1}}}

\newcommand{\dataset}{\mathcal{A}}

\newcommand{\contextenc}{f_\theta}

\newcommand{\alltasks}{\mathcal{W}}

\newcommand{\supportset}{\mathcal{S}}
\newcommand{\queryset}{\mathcal{Q}}

\newcommand{\bertclassifier}{BERT-classifier}
\newcommand{\bertknn}{BERT-kNN}

\renewcommand{\vec}[1]{\mathbf{#1}}

\newcommand{\cmark}{\ding{51}}
\newcommand{\xmark}{\ding{55}}

\newcommand\ti[1]{\textit{#1}}

\newcommand\tf[1]{\textbf{#1}}

\title{Non-Parametric Few-Shot Learning for Word Sense Disambiguation}

\author{
  Howard Chen, Mengzhou Xia, Danqi Chen \\
  \large{Princeton University} \\
  \texttt{\{howardchen, mengzhou, danqic\}@cs.princeton.edu} \\
}

\begin{document}
\maketitle

\begin{abstract}


Word sense disambiguation (WSD) is a long-standing problem in natural language processing.
One significant challenge in supervised all-words WSD is to classify among senses for a majority of words that lie in the long-tail distribution.
For instance, $84\%$ of the annotated words have less than $10$ examples in the SemCor training data.
This issue is more pronounced as the imbalance occurs in both word and sense distributions.
In this work, we propose $\ours$, a non-parametric few-shot learning approach to mitigate this data imbalance issue. By learning to compute distances among the senses of a given word through episodic training, $\ours$ transfers knowledge (a learned metric space) from high-frequency words to infrequent ones.
$\ours$ constructs the training episodes tailored to word frequencies and explicitly addresses the problem of the skewed distribution, as opposed to mixing all the words trained with parametric models in previous work.
Without resorting to any lexical resources, $\ours$ obtains strong performance against parametric alternatives, achieving a $75.1$ F1 score on the unified WSD evaluation benchmark~\cite{raganato2017uniwsd}. Our analysis further validates that infrequent words and senses enjoy significant improvement.\footnote{Our code is publicly available at: \url{https://github.com/princeton-nlp/metric-wsd}.}

\end{abstract}

\section{Introduction}

\label{sec:intro}

\begin{figure}[th]
\includegraphics[width=1.05\linewidth, height=5cm]{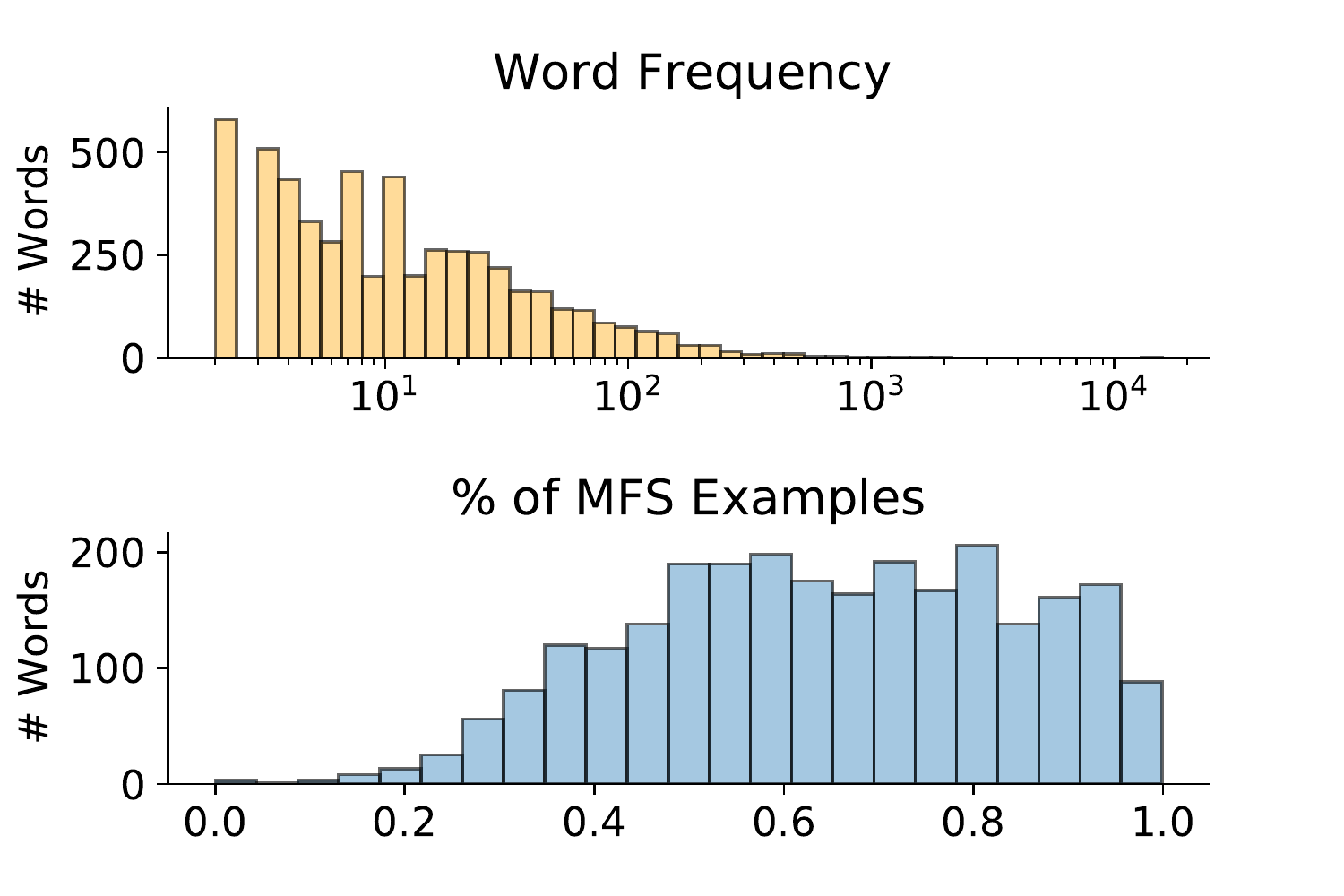}
\caption{Top: the long-tail distribution of words in the training data (SemCor).
Bottom: percentage of examples whose target senses are the most frequent sense for a given word (only words with $\geq 10$ examples are considered). All single-sense words are excluded.}
\label{fig:example}
\end{figure}

Word sense disambiguation (WSD) \citep{navigli2009word} is a widely studied problem that aims to assign words in text to their correct senses.
Despite advances over the years, a major challenge remains to be the naturally present data imbalance issue. Models suffer from extreme data imbalance, rendering learning the long-tail examples a major focus. In the English all-words WSD task \citep{raganato2017uniwsd}, $84\%$ of the annotated words\footnote{Here we use ``word'' for simplicity. In WSD datasets, a word is a combination of its stem and part-of-speech tag.} have less than $10$ occurrences in the training data and the most frequent sense (MFS) accounts for a large portion of the examples, resulting in a $65.2$ test F1 score by simply predicting MFS (Figure~\ref{fig:example}).

Recent approaches tackle this problem by resorting to extra sense information such as gloss (sense definition) and semantic relations to mitigate the issue of rare words and senses \citep{luo2018incorporating, luo2018leveraging, kumar2019zero, huang2019glossbert, blevins2020moving, bevilacqua2020breaking}. However, most work sticks to the parametric models that share parameters between words and adopts standard supervised learning mixing all the words of different frequencies. We argue that this accustomed paradigm exposes a missing opportunity to explicitly address the data imbalance issue.

In this work, we propose $\ours$, a simple non-parametric model coupled with episodic training to solve the long-tail problem, drawing inspiration from few-shot learning methods such as Prototypical Networks \citep{snell2017prototypical}.
Given a word, the model represents its senses by encoding a sampled subset (\textit{support set}) of the training data and learns a distance metric between these sense representations and the representations from the remaining subset (\textit{query set}).
This lightens the load for a model by learning an effective metric space
instead of learning a sense representation from scratch.
By sharing only the parameters in the text encoder, the model will trickle the knowledge of the learned metric space down from high-frequency words to infrequent ones.
We devise a sampling strategy that takes word and sense frequency into account and constructs support and query sets accordingly. In combination, this non-parametric approach naturally fits in the imbalanced few-shot problems, which is a more realistic setting when learning from a skewed data distribution as in WSD.

We evaluate $\ours$ on the unified WSD evaluation benchmark \citep{raganato2017uniwsd}, achieving a $75.1\%$ test F1 and outperforming parametric baselines using only the annotated sense supervision. A further breakdown analysis shows that the non-parametric model outperforms the parametric counterparts in low-frequency words and senses, validating the effectiveness of our approach.


\section{Related Work}
\label{sec:related}
\vspace{-3pt}
Word sense disambiguation has been studied extensively as a core task in natural language processing.
Early work computes relatedness through concept-gloss lexical overlap without supervision \citep{lesk1986, banerjee2003extended}. Later work designs features to build word-specific classifiers (\textit{word expert}) \citep{zhong2010makessense, shen2013corse,iacobacci2016embedding}.
All-words WSD unifies the datasets and training corpora by collecting large scale annotations \citep{raganato2017uniwsd}, which becomes the standard testbed for the WSD task.
However, due to the naturally present long-tail annotation, word expert approaches fall short in utilizing information across different words.

Recent supervised neural approaches prevail word-independent classifiers by more effective sentence feature extraction and achieve higher performance  \citep{kageback2016bilstm, raganato2017neural}. Approaches that use large pre-trained language models \cite{peters2018deep,devlin2019bert} further boost the performance \citep{hadiwinoto2019improved}. Recent work turns to incorporate gloss information \citep{luo2018incorporating, luo2018leveraging, huang2019glossbert, loureiro2019makessense, blevins2020moving}. Other work explores more lexical resources such as knowledge graph structures \citep{kumar2019zero, bevilacqua2020breaking, scarlini2020morecontext, scarlini2020contextenhanced}. All the above approaches mix words in the dataset and are trained under a standard supervised learning paradigm. Another close work to ours is \newcite{holla2020L2Lwsd}, which converts WSD into an $N$-way, $K$-shot few-shot learning problem and explores a range of meta-learning algorithms. This setup assumes disjoint sets of words between meta-training and meta-testing and deviates from the standard WSD setting.














%


\section{Method}
\label{sec:method}


\subsection{Task Definition}
\vspace{-4pt}
Given an input sentence $x=x_1, x_2, \ldots, x_n$, the goal of the \ti{all-words} WSD task is to assign a sense $y_i$ for every word $x_i$, where $y_i \in \mathcal{S}_{x_i} \subset \mathcal{S}$ for a given sense inventory such as the WordNet. In practice, not all the words in a sentence are annotated, and only a subset of positions are identified $\mathcal{I} \subseteq \{1, 2, \ldots, n\}$ to be disambiguated. 
The goal is to predict $y_i$ for $i \in \mathcal{I}$.

We regard all the instances of a word $w \in \mathcal{W}$ as a classification task $\mathcal{T}_w$, since only the instances of word $w$ share the output label set $\mathcal{S}_{w}$. We define input $\bar{x} = (x, t)$ where $x$ is an input sentence, and $1 \leq t \leq n$ is the position of the target word and the output is $y_t$ for $x_t$. A WSD system is a function $f$ such that $y = f(\bar{x})$. Our method groups the training instances by word $w$: $\mathcal{A}(w) = \{(\bar{x}^{(i)}, y^{(i)}): x^{(i)}_{t^{(i)}}=w\}_{i=1}^{\mathcal{N}(w)}$ where $\mathcal{N}(w)$ is the number of training instances for $\mathcal{T}_w$. It allows for word-based sampling as opposed to mixing all words in standard supervised training.

\subsection{Episodic Sampling}
We construct episodes by words with a tailored sampling strategy to account for the data imbalance issue. In each episode, all examples $\dataset(w)$ of a word $w$ are split into a support set $\supportset(w)$ containing $J$ distinct senses and a query set $\queryset(w)$ by a predefined ratio $r$ (splitting $r\%$ into the support set). When the support set is smaller than a predefined size $K$, we use the sets as they are. This split maintains the original sense distribution of the infrequent words as they will be used fully as support instances during inference. On the other hand, frequent words normally have abundant examples to form the support set. To mimic the few-shot behavior, we sample a \ti{balanced} number of examples per sense in the support set for frequent words (referred to as the $P_b$ strategy). We also compare to the strategy where the examples of all senses of the word are \ti{uniformly} sampled (referred to as the $P_u$ strategy).
We present the complete sampling strategy in Algorithm~\ref{episodic-sample}.

\begin{figure}[t]
\centering
\begin{minipage}{\linewidth}
\vspace{-10pt}
\begin{algorithm}[H]
\small
\caption{Episodic Sampling}
\label{episodic-sample}
\begin{algorithmic}[1]
\State $K$: maximum sample number for support set
\State $r$: support to query splitting ratio
\State $P$: sampling strategy $\in \{P_b, P_u\}$
\State Initialize empty dataset $\mathcal{D} = \emptyset$
\ForAll {$w \in \alltasks$}
    \State Retrieve $\dataset(w)$ and randomly split $\dataset(w)$ into $\tilde{\supportset}(w)$ and $\tilde{\queryset}(w)$ with a ratio $r$.

    \If {$|\tilde{\mathcal{S}}(w)| \leq K$}
    \State $\supportset(w) \leftarrow  \tilde{\supportset}(w)$; $\queryset(w) \leftarrow \tilde{\queryset}(w)$
    \Else
        \State $J \leftarrow$ \# of senses in $\tilde{\supportset}(w)$
        \State $\tilde{\mathcal{S}}_j(w) \leftarrow$ examples of sense $j$ in $\tilde{\supportset}(w)$

        \For {$k = 1 \dots |\tilde{\supportset}(w)|$}
            \State $j \leftarrow $ the sense of $k$-th example
            \State $\alpha_k \leftarrow
            \begin{cases}
                \frac{1}{|\tilde{\supportset_j}(w)| \times J}, & \text{if } P = P_b \text{ (balanced)}\\
                \frac{1}{|\tilde{\supportset}(w)|}, & \text{if } P = P_u \text{ (uniform)}
            \end{cases}$
        \EndFor
        \State ${\supportset}(w) \leftarrow \textsc{RandChoice}(\tilde{\supportset}(w), K, \alpha)$
        \State ${\queryset}(w) \leftarrow \tilde{\queryset}(w) \cup (\tilde{\supportset}(w) \setminus \supportset(w))$
    \EndIf
    \State $\mathcal{D} \leftarrow \mathcal{D} \cup \{ \supportset(w), \queryset(w) \}$
\EndFor \\
\Return $\mathcal{D}$
\end{algorithmic}
\end{algorithm}
\end{minipage}
\vspace{-10pt}
\end{figure}


\subsection{Learning Distance Metric}
We use BERT-base (uncased) \citep{devlin2019bert} as the context encoder.
We follow \citet{blevins2020moving} closely and denote context encoding as $\contextenc(\bar{x}) = \bert{BERT}(x)[t]$ where the context encoder is parameterized by $\theta$.  If a word $x_t$ is split into multiple word pieces, we take the average of their hidden representations.
In each episode, the model encodes the contexts in the support set $\supportset(w)$ and the query set $\queryset(w)$, where the encoded support examples will be taken average and treated as the sense representations (\textit{prototypes}).
For word $w$, the prototype
for sense $j$ among the sampled $J$ senses is computed from the support examples:
\begin{align*}
\vec{c}_j = \frac{1}{\lvert \supportset_j(w) \rvert} \sum_{(\bar{x}, y) \in \supportset_j(w)} \contextenc(\bar{x}), \numberthis \label{proto}
\end{align*}
where $\supportset_j(w) = \{(\bar{x}^{(i)}, y^{(i)}) : y^{(i)} = j\}_{i=1}^{\lvert \supportset_j \rvert} \subset \supportset(w)$.
We compute dot product\footnote{We experiment with negative squared $l_2$ distance as suggested in \newcite{snell2017prototypical} as the scoring function and find no improvement.} as the scoring function $s(\cdot, \cdot)$ between the prototypes and the query representations to obtain the probability of predicting sense $j$ given an example $(\bar{x}', y')$: 
\begin{align*}
p(y=j \mid \bar{x}') = \frac{\exp(s(\vec{c}_j, \contextenc(\bar{x}'))}{\sum_{k} \exp(s(\vec{c}_{k}, \contextenc(\bar{x}')))}. \numberthis \label{softmax}
\end{align*}
The loss is computed using negative log-likelihood
and is minimized through gradient descent. During inference, we randomly sample $\min(I_S, \lvert \dataset_j(w) \rvert)$ examples in the training set for sense $j$ as the support set, where $I_S$ is a hyperparameter. We also experimented with a cross-attention model which learns a scoring function for every pair of instances, similar to the BERT-pair model in \newcite{gao2019fewrel}; however, we didn't find it to perform better than the dual-encoder model.

\subsection{Relation to Prototypical Networks}
Our non-parametric approach is inspired and closely related to Prototypical Networks \citep{snell2017prototypical} with several key differences. First, instead of using disjoint tasks (i.e., words in our case) for training and testing, $\ours$ leverages the training data to construct the support set during inference. Second, we control how to sample the support set using a tailored sampling strategy (either balanced or uniform sense distribution).
This encourages learning an effective metric space from frequent examples to lower-frequency ones, which is different from adapting between disjoint tasks as in the typical meta-learning setup.




\begin{table*}[t]
\centering
\resizebox{1.0\columnwidth}{!}{%
\begin{tabular}{l c c cccc ccccc}
\toprule
   &  & \multicolumn{1}{c}{\textbf{Dev}} & \multicolumn{4}{c}{\textbf{Test Datasets}} & \multicolumn{5}{c}{\textbf{Concatenation of Test Datasets}} \\
    & Gloss? & SE07 & SE02 & SE03 & SE13 & SE15 & Nouns & Verbs & Adj. & Adv. & ALL \\
    \midrule
    WordNet S1 & \xmark & 55.2 & 66.8 & 66.2 & 63.0 & 67.8 & 67.6 & 50.3 & 74.3 & 80.9 & 65.2 \\
    Most frequent sense (MFS) & \xmark & 54.5 & 65.6 & 66.0 & 63.8 & 67.1 & 67.7 & 49.8 & 73.1 & 80.5 & 65.5 \\
    Bi-LSTM \cite{raganato2017neural} &  \xmark & 64.8 & 72.0 & 69.1 & 66.9 & 71.5 & 71.5 & 57.5 & 75.0 & 83.8 & 69.9 \\
    {\bertknn} & \xmark & 64.6 & 74.7 & 73.5 & 70.3 & 73.9 & 74.7 & 61.6 & 77.7 & 85.3 & 72.6\\
    {\bertclassifier} &  \xmark & 68.6 & 75.2 & 74.7 & 70.6 & 75.2 & 74.6 & 63.6 & 78.6 & \tf{87.0} & 73.5 \\
    1sent \cite{hadiwinoto2019improved} & \xmark & 67.0 & 75.0 & 71.6 & 69.7 & 74.4  & - & - & - & - & 72.7\\
    1sent+1sur$^\dagger$ \cite{hadiwinoto2019improved} & \xmark & 69.3 & 75.9 & 73.4 & 70.4 & 75.1 & - & - & - & -  & 73.7\\
\midrule
    \ours~(ours) & \xmark & \tf{71.4} & \tf{77.3} & \tf{75.6} & \tf{71.9} & \tf{76.6} & \tf{77.1} & \tf{64.9} & \tf{79.9} & 85.3 & \tf{75.1}\\
\midrule
    EWISE~\cite{kumar2019zero} & \cmark & 67.3 & 73.8 & 71.1 & 69.4 & 74.5 & 74.0 & 60.2 & 78.0 & 82.1 & 71.8 \\
    GlossBERT~\cite{huang2019glossbert} & \cmark & 72.5 & 77.7 & 75.2 & 76.1 & 80.4 & 79.8 & 67.1 & 79.6 & 87.4 & 77.0 \\
    EWISER~\cite{bevilacqua2020breaking} & \cmark & 71.0 & 78.9 & \tf{78.4} & 78.9 & 79.3 & \tf{81.7} & 66.3 & 81.2 & 85.8 & 78.3 \\
    BEM~\cite{blevins2020moving} & \cmark & \tf{74.5} & \tf{79.4} & {77.4} & \tf{79.7} & \tf{81.7} & {81.4} & \tf{68.5} & \tf{83.0} & \tf{87.9} & \tf{79.0} \\
\bottomrule
\end{tabular}
}
\caption{F1 scores for fine-grained all-words WSD task. We compare our non-parametric model against models without access to gloss information. We take results from EWISER where only supervised data and gloss are used but not WordNet examples. $\dagger$: surrounding sentences are used as extra context.
Our system uses the sampling strategy in Algorithm~\ref{episodic-sample} with $K=40, r=0.4, I_S=30, P=P_b$. 
}
\label{table:wsd_main}
\end{table*}


\begin{figure*}[th]
\includegraphics[trim={0 450 0 20},clip,width=0.97\linewidth]{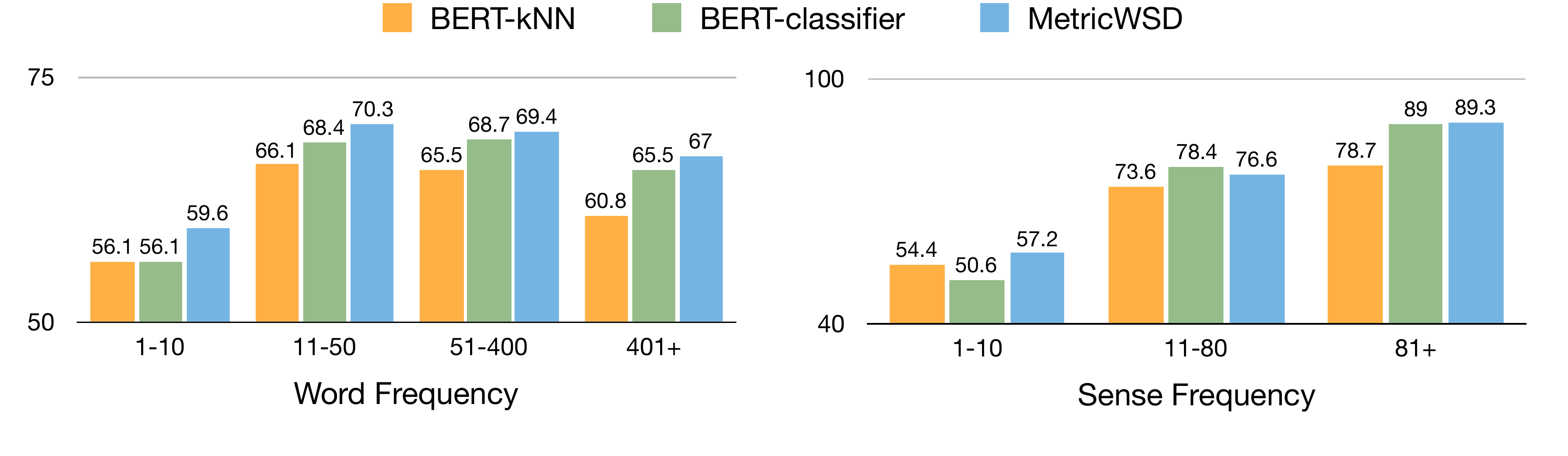}
\vspace{-0.7cm}
\caption{Accuracy breakdown by word frequency and sense frequency on the test set.
}
\label{fig:breakdown}
\end{figure*}
\vspace{-0.3cm}

\section{Experiments}
\label{sec:exp}
\vspace{-5pt}
We evaluate our approach with the WSD framework proposed by \newcite{raganato2017uniwsd}. We train our model on SemCor 3.0 and use SemEval-2007 (SE07) for development and the rest: Senseval-2 (SE02), Senseval-3 (SE03), SemEval-2013 (SE13), and SemEval-2015 (SE15) for testing. Following standard practice, we report performance on the separate test sets, the concatenation of all test sets, and the breakdown by part-of-speech tags. For all the experiments, we use the BERT-base (uncased) model as the text encoder.

\vspace{-0.5em}
\paragraph{Baselines}
We first compare to two simple baselines: {WordNet S1} always predicts the first sense and {MFS} always predicts the most frequent sense in the training data. We compare our approach to {\bertclassifier}: a linear classifier built on top of BERT (all the weights are learned together). As opposed to our non-parametric approach, the {\bertclassifier} has to learn the output weights from scratch.
We compare to another supervised baseline using contextualized word representations that extends the input context text with its surrounding sentences in the SemCor dataset  \citep{hadiwinoto2019improved}.
We also compare to a non-parametric nearest neighbor baseline {\bertknn}, which obtains sense representations by averaging BERT encoded representations from training examples of the same sense. It predicts the nearest neighbor of the input among the sense representations. The BERT weights are frozen which, different from our approach, does not learn the metric space.
Models using only supervised WSD data fall back to predicting the most frequent sense (MFS) when encountering unseen words.
For reference, we also list the results of recent state-of-the-art methods that incorporate gloss information including EWISE~\cite{kumar2019zero}, EWISER~\cite{bevilacqua2020breaking}, GlossBERT~\cite{huang2019glossbert}, and BEM~\cite{blevins2020moving}.
More implementation details are given in Appendix~\ref{sec:appendix}.





\begin{figure*}[!th]
\includegraphics[width=1.0\linewidth]{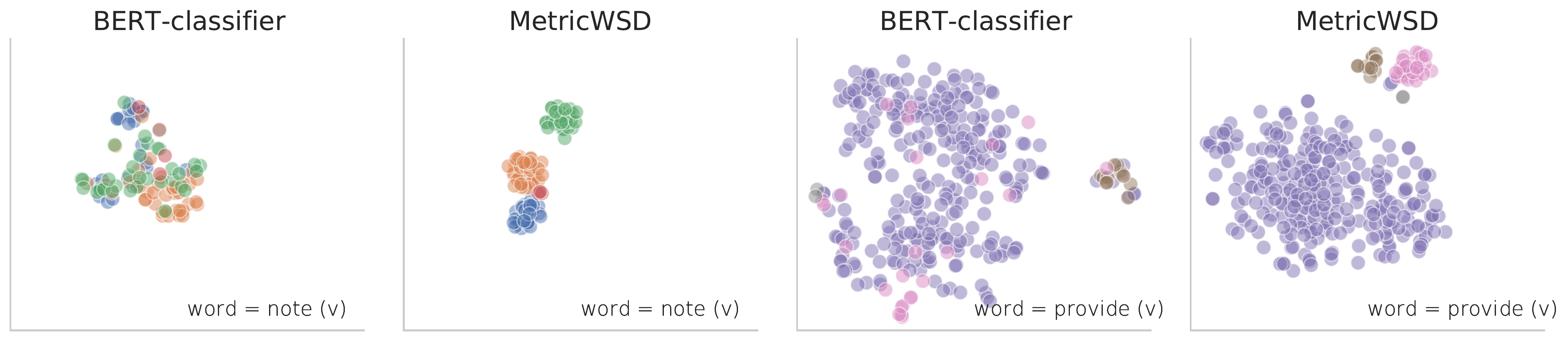}
\caption{t-SNE visualization of the learned representations $f_{\theta}(\bar{x})$ for the examples of $\ti{note}$ (v) and $\ti{provide}$ (v) in the SemCor dataset. It shows that {\ours} is better than {\bertclassifier} in grouping different senses. }
\label{fig:visualization}
\end{figure*}

\begin{table*}[th!]
\centering
\resizebox{0.98 \columnwidth}{!}{
\begin{tabular}{p{7.5cm}p{5.5cm}p{5.5cm}}
\toprule
\tf{Context} & \tf{{\bertclassifier} prediction} & \tf{$\ours$ prediction} \\
\midrule
\multirow{2}{*}{\parbox{7.5cm}{The \textbf{\textit{art}} of change-ringing is peculiar to the English, and, like most English peculiarities, unintelligible to the rest of the world.}} & art\%1:06:00:: (freq = 48) & art\%1:09:00:: (freq = 6) \\ \cmidrule(lr){2-2}\cmidrule(lr){3-3}
 &  the products of human creativity; works of art collectively & a superior skill that you can learn by study and practice and observation \\ \midrule
 \multirow{2}{*}{\parbox{7.5cm}{Eyes that were \textbf{\textit{clear}}, but also bring with a strange intensity, a sort of cold fire burning behind him.}} & clear\%3:00:00:: (freq = 45) & clear\%3:00:02:: (freq = 4) \\ \cmidrule(lr){2-2}\cmidrule(lr){3-3}
 & readily apparent to the mind  & allowing light to pass through \\ \midrule
  \multirow{2}{*}{\parbox{7.5cm}{And, according to \textbf{\textit{reports}} from US broadcaster CNBC, Citigroup is also planning to replay the state support.}} & report\%1:10:03:: (freq = 50) & report\%1:10:00:: (freq = 9) \\ \cmidrule(lr){2-2}\cmidrule(lr){3-3}
 & a written document describing the findings of some individual or group  & a short account of the news \\
\bottomrule
\end{tabular}
}
\vspace{-0.3em}
\caption{Examples of contexts and model predictions. The bold italic words in the contexts are disambiguated by {\bertclassifier} and $\ours$. We present the predicted sense key, the corresponding sense definition, and the sense frequency in the training set.}
\label{table:qualitative}
\end{table*}

\vspace{-0.5em}
\paragraph{Overall results}
Table~\ref{table:wsd_main} presents the overall results on the WSD datasets. Comparing against systems without using gloss information, $\ours$ achieves strong performance against all baselines. In particular, $\ours$ outperforms {\bertclassifier} by $1.4$ points and {\bertknn} by $2.5$ points respectively in F1 score on the test set. Using gloss information boosts the performance by a large margin especially for unseen words, where systems without access to gloss can only default to the first sense. We believe adding gloss has the potential to enhance the performance for our non-parametric approach and we leave it to future work.
\paragraph{Performance on infrequent words and senses}
The performance breakdown for words and senses of different frequency groups is given in Figure~\ref{fig:breakdown}. The non-parametric methods (both {\ours} and {\bertknn}) are better at handling infrequent words and senses. In particular, our approach outperforms {\bertclassifier} $3.5\%$ for the words with $\leq 10$ occurrences and $6.6\%$ for the senses with $\leq 10$ occurrences. It demonstrates the effectiveness of $\ours$ to handle scarce examples.
\paragraph{Ablation on sampling strategies}
We provide an ablation study for the sampling strategy on the development set. The system using the balanced strategy ($P_b$) achieves a $71.4$ F1 on the development set and drops to $69.2$ F1 when the uniform strategy ($P_u$) is used. Balancing the sampled senses achieves significantly higher performance than sampling with the uniform distribution and this observation is consistent across different hyper-parameter settings.







\section{Analysis}
\label{sec:analysis}



\vspace{-0.2em}
\paragraph{Qualitative analysis}
Table~\ref{table:qualitative} shows the examples which are correctly predicted by our method but incorrectly predicted by {\bertclassifier}. We see that $\ours$ is able to correctly predict the sense \textit{art\%1:09:00::} (a superior skill that you can learn by study and practice and observation), which has only $6$ training examples. The {\bertclassifier} model incorrectly predicts the sense \textit{art\%1:06:00::} (the  products  of  human  creativity; works of art collectively) that has many more training examples.

\vspace{-0.2em}
\paragraph{Visualization of learned representations}
We conduct a qualitative inspection of the learned representations for the {\bertclassifier} model and $\ours$. Figure~\ref{fig:visualization} shows the encoded representations of all $105$ examples in the SemCor dataset of the word \textit{note} (with part-of-speech tag \ti{v}). We see that the {\bertclassifier} model fails to learn distinct grouping of the senses while $\ours$ forms clear clusters. Note that even for the sense (red) with only few examples, our method is able to learn representations that are meaningfully grouped. Similarly, $\ours$ separates senses more clearly than {\bertclassifier} for the word \textit{provide} (with part-of-speech tag \ti{v}, especially on the rare sense (pink).



\section{Conclusion}
\vspace{-0.2em}
\label{sec:conclusion}
In this work, we introduce $\ours$, a few-shot non-parametric approach for solving the data imbalance issue in word sense disambiguation. Through learning the metric space and episodic training, the model learns to transfer knowledge from frequent words to infrequent ones. $\ours$ outperforms previous methods only using the standard annotated sense supervision and shows significant improvements on low-frequency words and senses. In the future, we plan to incorporate lexical information to further close the performance gap.

\vspace{-0.2em}
\section*{Acknowledgement}
\label{sec:ack}
We thank the members of the Princeton NLP group and the anonymous reviewers for their valuable comments and feedback. We also thank Terra Blevins at University of Washington for providing code and checkpoints for the baselines. Both HC and MX are supported by a Graduate Fellowship at Princeton University.

\newpage
\section*{Ethical Considerations}
\label{sec:ethics}

We identify areas where the WSD applications and our proposed approach will impact or benefit users. WSD systems are often used as an assistive submodule for other downstream tasks, rendering the risk of misuse less pronounced. However, it might still exhibit risk when biased data incurs erroneous disambiguation. For example, the word ``shoot'' might have a higher chance to be interpreted as a harmful action among other possible meanings when the context contains certain racial or ethnic groups that are biasedly presented in training data. Our proposed method does not directly address this issue. Nonetheless, we identify the opportunity for our approach to alleviate the risk by providing an easier way to inspect and remove biased prototypes instead of making prediction using learned output weights that are hard to attribute system's biased behavior. We hope future work extends the approach and tackles the above problem more explicitly.

\bibliography{ref}

\begin{thebibliography}{24}
\expandafter\ifx\csname natexlab\endcsname\relax\def\natexlab#1{#1}\fi

\bibitem[{Banerjee and Pedersen(2003)}]{banerjee2003extended}
Satanjeev Banerjee and Ted Pedersen. 2003.
\newblock Extended gloss overlaps as a measure of semantic relatedness.
\newblock In \emph{International Joint Conference on Artificial Intelligence
  (IJCAI)}.

\bibitem[{Bevilacqua and Navigli(2020)}]{bevilacqua2020breaking}
Michele Bevilacqua and Roberto Navigli. 2020.
\newblock Breaking through the 80\% glass ceiling: Raising the state of the art
  in word sense disambiguation by incorporating knowledge graph information.
\newblock In \emph{Association for Computational Linguistics (ACL)}, pages
  2854--2864.

\bibitem[{Blevins and Zettlemoyer(2020)}]{blevins2020moving}
Terra Blevins and Luke Zettlemoyer. 2020.
\newblock {M}oving down the long tail of word sense disambiguation with gloss
  informed bi-encoders.
\newblock In \emph{Association for Computational Linguistics (ACL)}.

\bibitem[{Devlin et~al.(2019)Devlin, Chang, Lee, and
  Toutanova}]{devlin2019bert}
Jacob Devlin, Ming-Wei Chang, Kenton Lee, and Kristina Toutanova. 2019.
\newblock {BERT}: Pre-training of deep bidirectional transformers for language
  understanding.
\newblock In \emph{North American Association for Computational Linguistics
  (NAACL)}, pages 4171--4186.

\bibitem[{Gao et~al.(2019)Gao, Han, Zhu, Liu1, Li, Sun, and
  Zhou}]{gao2019fewrel}
Tianyu Gao, Xu~Han, Hao Zhu, Zhiyuan Liu1, Peng Li, Maosong Sun, and Jie Zhou.
  2019.
\newblock Fewrel 2.0: Towards more challenging few-shot relation
  classification.
\newblock In \emph{Empirical Methods in Natural Language Processing (EMNLP)}.

\bibitem[{Hadiwinoto et~al.(2019)Hadiwinoto, Ng, and
  Gan}]{hadiwinoto2019improved}
Christian Hadiwinoto, Hwee~Tou Ng, and Wee~Chung Gan. 2019.
\newblock Improved word sense disambiguation using pre-trained contextualized
  word representations.
\newblock In \emph{Empirical Methods in Natural Language Processing (EMNLP)},
  pages 5297--5306.

\bibitem[{Holla et~al.(2020)Holla, Mishra, Yannakoudakis, and
  Shutova}]{holla2020L2Lwsd}
Nithin Holla, Pushkar Mishra, Helen Yannakoudakis, and Ekaterina Shutova. 2020.
\newblock Learning to learn to disambiguate: Meta-learning for few-shot word
  sense disambiguation.
\newblock In \emph{Findings of the Empirical Methods in Natural Language
  Processing (EMNLP Findings)}.

\bibitem[{Huang et~al.(2019)Huang, Sun, Qiu, and Huang}]{huang2019glossbert}
Luyao Huang, Chi Sun, Xipeng Qiu, and Xuan-Jing Huang. 2019.
\newblock {GlossBERT}: {BERT} for word sense disambiguation with gloss
  knowledge.
\newblock In \emph{Empirical Methods in Natural Language Processing (EMNLP)},
  pages 3500--3505.

\bibitem[{Iacobacci et~al.(2016)Iacobacci, Pilehvar, and
  Navigli}]{iacobacci2016embedding}
Ignacio Iacobacci, Mohammad~Taher Pilehvar, and Roberto Navigli. 2016.
\newblock Embeddings for word sense disambiguation: An evaluation study.
\newblock In \emph{Association for Computational Linguistics (ACL)}.

\bibitem[{Kumar et~al.(2019)Kumar, Jat, Saxena, and Talukdar}]{kumar2019zero}
Sawan Kumar, Sharmistha Jat, Karan Saxena, and Partha Talukdar. 2019.
\newblock Zero-shot word sense disambiguation using sense definition
  embeddings.
\newblock In \emph{Association for Computational Linguistics (ACL)}, pages
  5670--5681.

\bibitem[{Kågebäck and Salomonsson(2016)}]{kageback2016bilstm}
Mikael Kågebäck and Hans Salomonsson. 2016.
\newblock Word sense disambiguation using a bidirectional lstm.
\newblock In \emph{International Conference on Computational Linguistics
  (COLING)}.

\bibitem[{Lesk(1986)}]{lesk1986}
Michael Lesk. 1986.
\newblock Automatic sense disambiguation using machine readable dictionaries:
  How to tell a pine cone from an ice cream cone.
\newblock In \emph{Conference on Systems Documentation (SIGDOC)}.

\bibitem[{Loureiro and Jorge(2019)}]{loureiro2019makessense}
Daniel Loureiro and Alípio Jorge. 2019.
\newblock Language modelling makes sense: Propagating representations through
  wordnet for full-coverage word sense disambiguation.
\newblock In \emph{Association for Computational Linguistics (ACL)}.

\bibitem[{Luo et~al.(2018{\natexlab{a}})Luo, Liu, He, Xia, Sui, and
  Chang}]{luo2018leveraging}
Fuli Luo, Tianyu Liu, Zexue He, Qiaolin Xia, Zhifang Sui, and Baobao Chang.
  2018{\natexlab{a}}.
\newblock Leveraging gloss knowledge in neural word sense disambiguation by
  hierarchical co-attention.
\newblock In \emph{Empirical Methods in Natural Language Processing (EMNLP)},
  pages 1402--1411.

\bibitem[{Luo et~al.(2018{\natexlab{b}})Luo, Liu, Xia, Chang, and
  Sui}]{luo2018incorporating}
Fuli Luo, Tianyu Liu, Qiaolin Xia, Baobao Chang, and Zhifang Sui.
  2018{\natexlab{b}}.
\newblock Incorporating glosses into neural word sense disambiguation.
\newblock In \emph{Association for Computational Linguistics (ACL)}, pages
  2473--2482.

\bibitem[{Navigli(2009)}]{navigli2009word}
Roberto Navigli. 2009.
\newblock Word sense disambiguation: A survey.
\newblock \emph{ACM computing surveys (CSUR)}, 41(2):1--69.

\bibitem[{Peters et~al.(2018)Peters, Neumann, Iyyer, Gardner, Clark, Lee, and
  Zettlemoyer}]{peters2018deep}
Matthew Peters, Mark Neumann, Mohit Iyyer, Matt Gardner, Christopher Clark,
  Kenton Lee, and Luke Zettlemoyer. 2018.
\newblock Deep contextualized word representations.
\newblock In \emph{North American Association for Computational Linguistics
  (NAACL)}, volume~1, pages 2227--2237.

\bibitem[{Raganato et~al.(2017{\natexlab{a}})Raganato, Bovi, and
  Navigli}]{raganato2017neural}
Alessandro Raganato, Claudio~Delli Bovi, and Roberto Navigli.
  2017{\natexlab{a}}.
\newblock Neural sequence learning models for word sense disambiguation.
\newblock In \emph{Empirical Methods in Natural Language Processing (EMNLP)},
  pages 1156--1167.

\bibitem[{Raganato et~al.(2017{\natexlab{b}})Raganato, Camacho-Collados, and
  Navigli}]{raganato2017uniwsd}
Alessandro Raganato, Jose Camacho-Collados, and Roberto Navigli.
  2017{\natexlab{b}}.
\newblock Word sense disambiguation: A unified evaluation framework and
  empirical comparison.
\newblock In \emph{European Chapter of the Association for Computational
  Linguistics (EACL)}.

\bibitem[{Scarlini et~al.(2020{\natexlab{a}})Scarlini, Pasini, and
  Navigli}]{scarlini2020contextenhanced}
Bianca Scarlini, Tommaso Pasini, and Roberto Navigli. 2020{\natexlab{a}}.
\newblock Sensembert: Context-enhanced sense embeddings for multilingual word
  sense disambiguation.
\newblock In \emph{Conference on Artificial Intelligence (AAAI)}.

\bibitem[{Scarlini et~al.(2020{\natexlab{b}})Scarlini, Pasini, and
  Navigli}]{scarlini2020morecontext}
Bianca Scarlini, Tommaso Pasini, and Roberto Navigli. 2020{\natexlab{b}}.
\newblock With more contexts comes better performance: Contextualized sense
  embeddings for all-round word sense disambiguation.
\newblock In \emph{Empirical Methods in Natural Language Processing (EMNLP)}.

\bibitem[{Shen et~al.(2013)Shen, Bunescu, and Mihalcea}]{shen2013corse}
Hui Shen, Razvan Bunescu, and Rada Mihalcea. 2013.
\newblock Coarse to fine grained sense disambiguation in wikipedia.
\newblock In \emph{Conference on Lexical and Computational Semantics (SEM)}.

\bibitem[{Snell et~al.(2017)Snell, Swersky, and Zemel}]{snell2017prototypical}
Jake Snell, Kevin Swersky, and Richard Zemel. 2017.
\newblock Prototypical networks for few-shot learning.
\newblock In \emph{Advances in Neural Information Processing Systems (NIPS)},
  pages 4077--4087.

\bibitem[{Zhong and Ng(2010)}]{zhong2010makessense}
Zhi Zhong and Hwee~Tou Ng. 2010.
\newblock It makes sense: A wide-coverage word sense disambiguation system for
  free text.
\newblock In \emph{Association for Computational Linguistics (ACL)}.

\end{thebibliography}
\bibliographystyle{acl_natbib}
\newpage
\clearpage

\appendix
\section{Appendix}
\label{sec:appendix}

\subsection*{Implementation Details}
When constructing support and query sets, we skip words that contain only one sense or words that have only one example per sense since they cannot be split into meaningful support and query sets (e.g., only two senses and one example each for a word). During inference, all the aforementioned examples are used as supports.

For {\bertclassifier}, all parameters are fine-tuned during supervised training. We use batch size of $4$ sentences, and runs for $20$ epochs. Our non-parametric model constructs an episode for each word type, accumulate gradients for $5$ episodes before performing gradient update, runs for $100$ epochs. All models use the AdamW optimizer with learning rate 1e-5.

\end{document}